# Actor-Critic Network for Q&A in an Adversarial Environment

**Bejan Sadeghian,** University of Texas at Austin
bejan.sadeghian@utexas.edu

## Abstract

Significant work has been placed in the Q&A NLP space to build models that are more robust to adversarial attacks. Two key areas of focus are in generating adversarial data for the purposes of training against these situations or modifying existing architectures to build robustness within. This paper introduces an approach that joins these two ideas together to train a critic model for use in an almost reinforcement learning framework. Using the Adversarial SQuAD "Add One Sent" dataset we show that there are some promising signs for this method in protecting against Adversarial attacks.

## 1 Introduction

Within Reinforcement Learning, Actor-Critic frameworks are commonly used to inform one model, the actor, if their decision was a positive or negative one based on the second model's assessment (the critic). This two model relationship is generally used in the training loop in situations where a wrong decision is costly and there is limited or no training data for each situation. These ideas are also popular in GANs for data generation.

In the case of adversarial attacks, the intention is to trick the actor into making a wrong decision. To offset this, a critic model can help by checking if each decision that would be made is a good or a bad one and inform the actor so the decision can be modified if needed.

In this paper we test the Actor-Critic approach out of the box but identify some limitations when including it in one training loop. We then recommend an alternative that shows some promise by using a separately trained critic in the actor's training loop for guidance.

In addition to using an Actor-Critic framework to we took to automatically generate adversarial data. This was required to properly teach the critic network how to discriminate between good and bad. We borrow ideas of permuting the input data from Adversarial Examples for Evaluating Reading Comprehension Systems (AEERCS) (Jia and Liang, 2017) to generate this data.

For all parts of this paper we focused on keeping the actor model as similar to the baseline version as possible and looked to either change the training data, the training loop, or the inference decisions. This allows us to make a strong comparison to the baseline model provided.

## 2 Generating Adversarial Data

To generate adversarial data that could fool our actor model, we tested two approaches. In all cases we would only modify the "passage" not the "question".

Our first approach involved permuting the input embedding by masking or zeroing out values at random to generate a negative sample. The randomness here was tested at different levels of replacement/dropout probability and tested between only masking non-stop words or all words.

Our second approach borrows from Adversarial Example Generation with Syntactically Controlled Paraphrase Network (AEGSCPN) (Iyyer et al., 2018) and AEERCS (Jia and Liang, 2017). The authors here modify the input text data based either on the parse tree to create syntactically accurate examples or on the part of speech in the case of the 2017 paper. Both of these approaches showed that this adversary dataset would trick high performing Q&A models.

Ultimately we chose to pursue the second approach but with ideas borrowed from the first for computational reasons. Instead of basing replacement on the back calculated parse tree, we chose to replace words in a negative sample at random. In this case, replacements in the passage were taken from the question. The intent here was to train the model to not be fooled by "similar enough words to the question" which we know self-attention models can tend to do.

| Question |
|---|
| what is the grotto at notre dame ? |
| **Golden Passage Span** |
| a marian place of prayer |
| **Generated Negative Passage Span** |
| a is the place at prayer dame |

Table 1: Adversarial Example (50% replacement)

The actor model used the question and passage from the publically available SQuAD dataset for training. However this was not the case for the critic model. Because the critic was trained to identify anomalies in the passage text. It used the *text preceding the golden passage span* as its query and either the golden passage span or the negative generated passage span. Figure 1 shows an example of this, notice the difference in the positive and negative sample's span after the first [SEP]. The positive instance is the golden span, the negative instance is a random mixture of the question and golden. The goal here was to identify adversarial spans *within* the passage text.

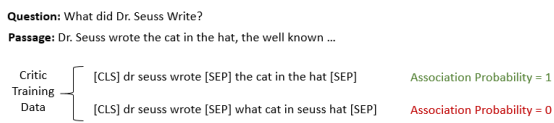

Figure 1: Critic Training Data w/ positive and negative observations

One important item to note here, this generated adversarial data is only ever seen by our critic model. The critic never sees or tries to infer from the question so there was no risk of leaking information from the question into the answer during its training. And since the training data for this model never gets exposed to the actor's there was no risk of information leakage there as well. Examples of this data are shown in Table 1.

## 3 Model Architectures

Different methods were tested during this research to identify an effective actor-critic network. A key consideration here was to maintain a non-computationally expensive approach since we were using two different embeddings for each and translation between the two was costly.

### 3.1 Actor Model

The Actor model architecture, an Encoder network for predicting start and end positions of each answer, was left mostly untouched in our experiments. This model was provided by the UTCS department as a baseline to compare against[1]. This was done to focus improvements of the critic model and the actor-critic network as opposed to making improvements on the actor model itself. While the architecture remained the same, we did apply small modifications around this model for either adjusting the loss during training or changing its decision during inference.

### 3.2 Critic Model

Our experiments involved training a critic model both alongside the actor as well as separately. In a true Actor-Critic framework losses are communicated to the actor from the critic. We did not deviate from this framework.

#### 3.2.1 Encoder-Decoder with Attention

The first architecture tested for our critic network was an Encoder-Decoder LSTM network with Attention (Loung et al., 2015). This architecture has historically proven to be effective in sequence to sequence translation.

Our goal with this architecture was to have the query text (text preceding the golden span) be the input of the encoder and the passage span be the input of the decoder (along with the hidden state from the encoder). Having the decoder's output attend to the encoder's output allowed for a relationship representation between the query and passage span text. These outputs were then passed into a dense network where a prediction would be made as to if the query text and the passage span were associated or not.

#### 3.2.2 BERT with a Dense Classification Head

The second architecture we tested used the pre-trained BERT model (Devlin et al., 2019) with a custom three layer dense network as it's classification head. BERT has proven to be one of the most performant models for many NLP tasks. This is particularly true for text sequence based classification which is our use case.

---

[1] https://github.com/gregdurrett/nlp-qa-finalproj#training-and-evaluation

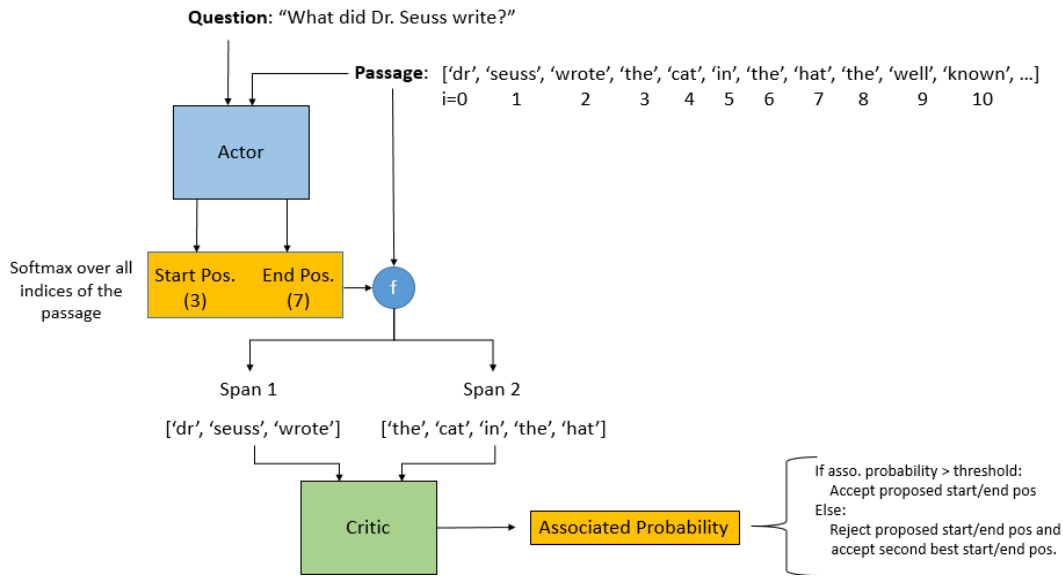

Figure 2: Actor-Critic Network Diagram

Because BERT was pre-trained, to modify it slightly we unfroze it's parameters for a single epoch in the middle of training. This allowed for some fine tuning at a very low learning rate. This was only done after a few epochs to allow the dense network to first form so proper gradients could be used.

### 3.3 Actor – Critic Framework

Figure 2 depicts how the actor and critic models worked together. Originally we had begun by training both networks simultaneously; however, due to different rates of learning and poor predictions by the critic early on we chose to separate the two.

This meant we trained the critic first in its own training loop then used the pre-trained model in the actor's training loop to correct the actor's predictions. This was done by attaching the critic model to the actor's forward method and adding the loss from the critic's prediction along with the actor's prediction loss.

To describe the order of operations fully. We first trained a critic network to predict if a span was abnormal/adversarial based on that span's preceding text. Once we had that model trained and frozen, we then trained the actor to predict the start and end positions of the span in the passage based on the question presented. Before committing to that answer we would let the critic assign a probability that the proposed span (slicing the passage according to the start/end positions) and the preceding passage text were abnormal or not. If the probability was less than our threshold it would indicate that the two texts were not associated and the critic would therefore reject the actor's original proposal. The 'argmax' across the actor's output out then be taken again with the exclusions removed and we would allow those proposed indices be the actor's prediction.

The loss function during training of the actor averaged the Cross Entropy loss (for start and end positions) and the Binary Cross Entropy loss (the critic's Boolean prediction) together in equal parts. This incentivized the actor to more strongly weigh the observations that the critic was not very confident on (if it was much lower than "1") and place less weight on the ones the critic was very confident in. Because we continued to include the loss of the start and end positions the actor would always update towards predicting the correct start and end positions.

## 4  Results and Discussion

There are two sets of results to share from our experiment. The first is specific to the performance of the critic model when trained separately. The second is on the ultimate goal which is to identify and correctly change decisions when an adversarial passage fools the actor model.

## 4.1 Establishing a Baseline

In order to establish a baseline to compare against, we first trained the baseline (Actor) model with no modification or critic on the SQuAD dataset. We were able to match similar performance numbers to its reference point[2]. These parameters were then frozen for all testing going forward.

## 4.2 Critic Classification Results

We then tested both the BERT and Encoder-Decoder network's performance on the Sequence classification problem to identify adversarial text in our passage. Performance of is shown in Table 2. Overall, the BERT model performed best.

The two probability values in the results are from different levels of replacement probability when forming the adversarial data (larger number means greater probability of using the question text).

| Model | Replacement Probability | Accuracy |
|---|---|---|
| Encoder-Decoder | 75 | 56.5% |
| | 90 | 59.3% |
| BERT | 75 | 61.1% |
| | 90 | 65.3% |

Table 2: Critic Performance.

Our next task was to select a threshold for our critic to override the actor's decision. Because anomaly detection needs to be wary of the high cost of making a false positive (i.e. identifying an anomaly when it is not one) we set the threshold to flag an anomaly very low at p=0.3. Figure 3 shows why this threshold was selected.

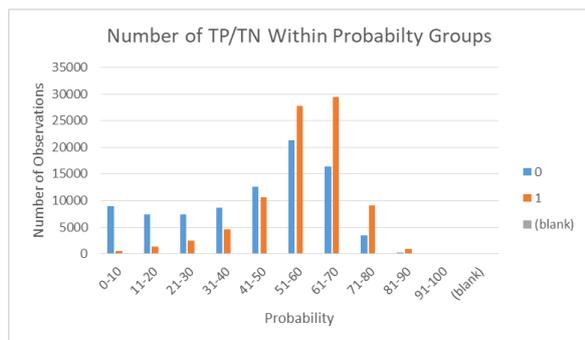

Figure 3: Probability Group Membership by Class

## 4.3 Adversarial Q&A Results

The ultimate goal of our work was to improve Q&A performance from our baseline using an Actor-Critic framework. Table 3 shows the results from our various experiments.

Results were mixed, while F1 scores tended to be equal to or higher with a critic in place, Exact match (EM) scores were lower than baseline in all cases. This suggests that while the critic may reject a starting or ending position correctly in some cases, the actor would still not correctly predict an accurate start *and* end position as its second guess.

| Model | F1 Score | EM Score |
|---|---|---|
| Baseline (No Critic) | 47.98 | 38.11 |
| Encoder - Decoder | 47.70 | 37.33 |
| BERT | 49.02 | 38.05 |
| | | |
| Non-Adversarial Baseline | 60.57 | 48.56 |

Table 3: Q&A Results.

An example from question ID 571c7abfdd7acb1400e4c0bb-high-conf-turk3 shows this behavior. For this question, *"What did Philo incorrectly assume that the air became?"* the actor's original prediction would have answered *"farnsworth incorrectly assumed that the air became something"* based on the adversarial sentence *"Farnsworth incorrectly assumed that the air became something else."* In this case, the critic correctly flagged the original predicted span as adversarial and rejected the actor's initial answer. However the actor's second best guess was not correct either, *"surmised that parts of the air in the vessel were converted into the classical element"*. Despite this miss it is a promising sign that the adversarial span was correctly flagged and the next best answer was closer to the real answer "fire" (the next word after "classical element").

## 5 Areas of Further Study

There are a few areas of improvement we could see being done on this research. The first is related to the use of the critic model's information. The second is related to how a rejection is handled. Third would be to look at the actor itself. And finally, improvements could be made on how adversarial data was generated. We will explore all three in more detail.

## 5.1 Critic Model's Information

One of the largest barriers we faced when making a classification model based on text is the fact that the input data varied in length with each observation. To get around this issue we simply used the pooling layer from the transformer model or in the case of the Encoder-Decoder averaged along the sentence dimension before inputting into the dense network. In both cases a lot of information is lost and this could be highly detrimental to the performance of the model.

The best alternative we can think of to get around this varying length problem while also maintaining most of the available information is to pad to a longer length than we would expect an input sentence to reach. Padding of course has its own issues, especially with the Encoder-Decoder network, but there is the potential for using padding with a 1D convolution network so all of the padding does not group at one side.

In addition, we mostly ignored the information coming from the transformer's hidden layers in our experiment. There is the potential to combine or replace the pooling layer's information with information from the hidden layers. Avoiding or supplementing the pooling layer's compression could prove fruitful.

## 5.2 Rejection Method

In our experiment, we allowed the critic to accept or reject the actor's decision on where each start/end index was. This rejection was a very basic decision to remove the specific index from being an option when calculating the "next best" start and end indices.

One alternative we thought to test was to remove all indices between the start and end positions proposed by the actor. This makes sense logical sense because if the critic is rejecting a start/end index based on the span between, we may want to reject the entire span from candidacy.

In addition to our rejection method, the fact that we only reject once is a limitation of our experiment and only fits when there is one adversarial span present. In the real world case we would not want to limit the number of rejections by a preselected value.

## 5.3 Actor Model

Throughout our experimentation we did not modify the actor architecture. This was by design so we could compare the effectiveness of the Actor-Critic network itself. However, to achieve a more performant pair of models one would reasonably want to look at how to improve both the critic and the actor.

One popular idea would be to plug BERT or another transformer in as the actor model here. Given the fact that transformers are very applicable to the Q&A problem it would make sense to apply this model type. The added benefit here is if both models are using the same architecture, or at least the same mapped embedding then token to ID translation would be much faster. In our case we modified the data generator to provide the raw token for each passage so that our critic model could encode the tokens as well. Originally the data generator only provided encoded tokens for the actor model.

## 5.4 Adversarial Data Generation

Our adversarial data generation was predominantly modeled on the prior knowledge that Q&A models are easily fooled by unrelated sentences in the passage that were closely matched to the question. For that reason we chose to replace at random parts of the golden span with the question to generate a negative span. This randomization was done after testing a few ideas like replacing only non-stop words but it was not pursued heavily.

As a follow up, we would like to explore the impact of more intelligently generating adversarial data using parts of speech. In particular, replacing nouns with other noun and verbs with other verbs. These replacements do not necessarily need to be sourced from the question but could be sourced from more general dictionaries too (e.g. replace "super" in "super bowl" with "mega"). This is an entire body of work which we reference in the related materials but could be the key to improving the critic's accuracy in identifying adversaries.

## 6 Conclusion

Our experiment shows us that actor-critic networks could be used similarly to how they are used in Reinforcing Learning. In fact, these models can work together not only during training, which is more common in RL, but also during inference to "catch" an error. Ideally the critic is used primarily during training to build a robust actor because of computational limits during inference (e.g. driving a car doesn't given enough time for both an actor and a critic to make decisions). But having a second filter to "second guess" the actor is not a bad thing.

While our model results show a slight improvement for the F1 score over baseline, the EM results showed lower performance across the board. We propose multiple improvement ideas to consider as well as the active research being done to build more robust models.